\documentclass[conference]{IEEEtran}
\IEEEoverridecommandlockouts
\usepackage{cite}
\usepackage{amsmath,amssymb,amsfonts}
\usepackage[pagebackref,breaklinks,colorlinks,bookmarks=false]{hyperref}
\usepackage{multirow}
\usepackage{booktabs}
\usepackage{algorithmic}
\usepackage{graphicx}
\usepackage{textcomp}
\usepackage{xcolor}
\def\BibTeX{{\rm B\kern-.05em{\sc i\kern-.025em b}\kern-.08em
    T\kern-.1667em\lower.7ex\hbox{E}\kern-.125emX}}
\begin{document}

\title{SC-HVPPNet: Spatial and Channel Hybrid-Attention Video Post-Processing Network with CNN and Transformer\\
\thanks{* Corresponding author.}
}

\author{\IEEEauthorblockN{1\textsuperscript{st} Tong Zhang}
\IEEEauthorblockA{\textit{School of Computer Science and Technology} \\
\textit{Harbin Institute of Technology}\\
Harbin, China \\
tongzhang@stu.hit.edu.cn} \\

\IEEEauthorblockN{3\textsuperscript{rd} Shaohui Liu}
\IEEEauthorblockA{\textit{School of Computer Science and Technology} \\
\textit{Harbin Institute of Technology}\\
Harbin, China \\
shliu@hit.edu.cn}
\and
\IEEEauthorblockN{2\textsuperscript{nd} Wenxue Cui}
\IEEEauthorblockA{\textit{School of Computer Science and Technology} \\
\textit{Harbin Institute of Technology}\\
Harbin, China \\
wxcui@hit.edu.cn} \\

\IEEEauthorblockN{4\textsuperscript{th} Feng Jiang*}
\IEEEauthorblockA{\textit{School of Computer Science and Technology} \\
\textit{Harbin Institute of Technology}\\
Harbin, China \\
fjiang@hit.edu.cn}
}
\maketitle

\begin{abstract}
Convolutional Neural Network (CNN) and Transformer have attracted much attention recently for video post-processing (VPP). However, the interaction between CNN and Transformer in existing VPP methods is not fully explored, leading to inefficient communication between the local and global extracted features. 
In this paper, we explore the interaction between CNN and Transformer in the task of VPP, and propose a novel Spatial and Channel Hybrid-Attention Video Post-Processing Network (SC-HVPPNet), which can cooperatively exploit the image priors in both spatial and channel domains. Specifically, in the spatial domain, a novel spatial attention fusion module is designed, in which two attention weights are generated to fuse the local and global representations collaboratively. In the channel domain, a novel channel attention fusion module is developed, which can blend the deep representations at the channel dimension dynamically. Extensive experiments show that SC-HVPPNet notably boosts video restoration quality, with average bitrate savings of 5.29\%, 12.42\%, and 13.09\% for Y, U, and V components in the VTM-11.0-NNVC RA configuration.
\end{abstract}

\begin{IEEEkeywords}
Video compression, CNN, Transformer, VVC, Post-processing, Quality enhancement
\end{IEEEkeywords}

\section{Introduction}
With the boom in Ultra High Definition (UHD) video, there is a growing demand for video services with higher resolution and lower data storage requirements. Efficient video compression techniques have also become one of the current hot topics within the field of computer vision. Accordingly, several new video coding standards are released for video compression and transmission, including Versatile Video Coding (VVC) ~\cite{VVC}, High Efficiency Video Coding (HEVC) ~\cite{6316136}, Advanced Video Coding (AVC) ~\cite{Wiegand2003Overview} and Elementary Video Coding (EVC) ~\cite{9146794}. Compared to HEVC/H.265, VVC/H.266 better supports high spatial resolution and high dynamic range. Considering the benefits of VVC, many novel coding techniques are integrated into the latest VVC standards, which greatly improves coding performance.

For the sake of improving video reconstruction quality and reducing noticeable visual artifacts, CNN and Transformer are widely proposed.
Recently, many video post-processing methods ~\cite{10008797, Liang2021SwinIRIR} are proposed, which can be roughly categorized into the following three groups: CNN-based methods, Transformer-based methods, CNN and Transformer fusion-based methods. 
Specifically, \textbf{1) CNN}-based methods aim to improve the video compression performance of existing coding modules.
Among them, some algorithms~\cite{9118561, 9897324} focus on residual learning and dense connections, which leads to a considerable coding performance.
Furthermore, to concentrate on the issue of content adaptation, a series of algorithms~\cite{10202562, 10018046} are proposed, which alleviates the gap between training and test data.
However, limited to the size of the convolution kernel, the receptive field of CNN makes it difficult to model global information, which limits its performance.
\textbf{2) Transformer}-based methods focus on the global interaction by the self-attention mechanism, which achieves promising performance in computer vision tasks~\cite{Yan2023ReferredBM}. 
Specifically, Swin Transformer~\cite{Liu2021SwinTH} is introduced into video projects so as to promote compression performance to a certain extent.
Nevertheless, Transformer also faces the obstacle of being insensitive to position when calculating global attention and ignoring details of local features.
To cope with the above issues, 
\begin{figure*}
  \centering
    \includegraphics[width=1.0\linewidth]{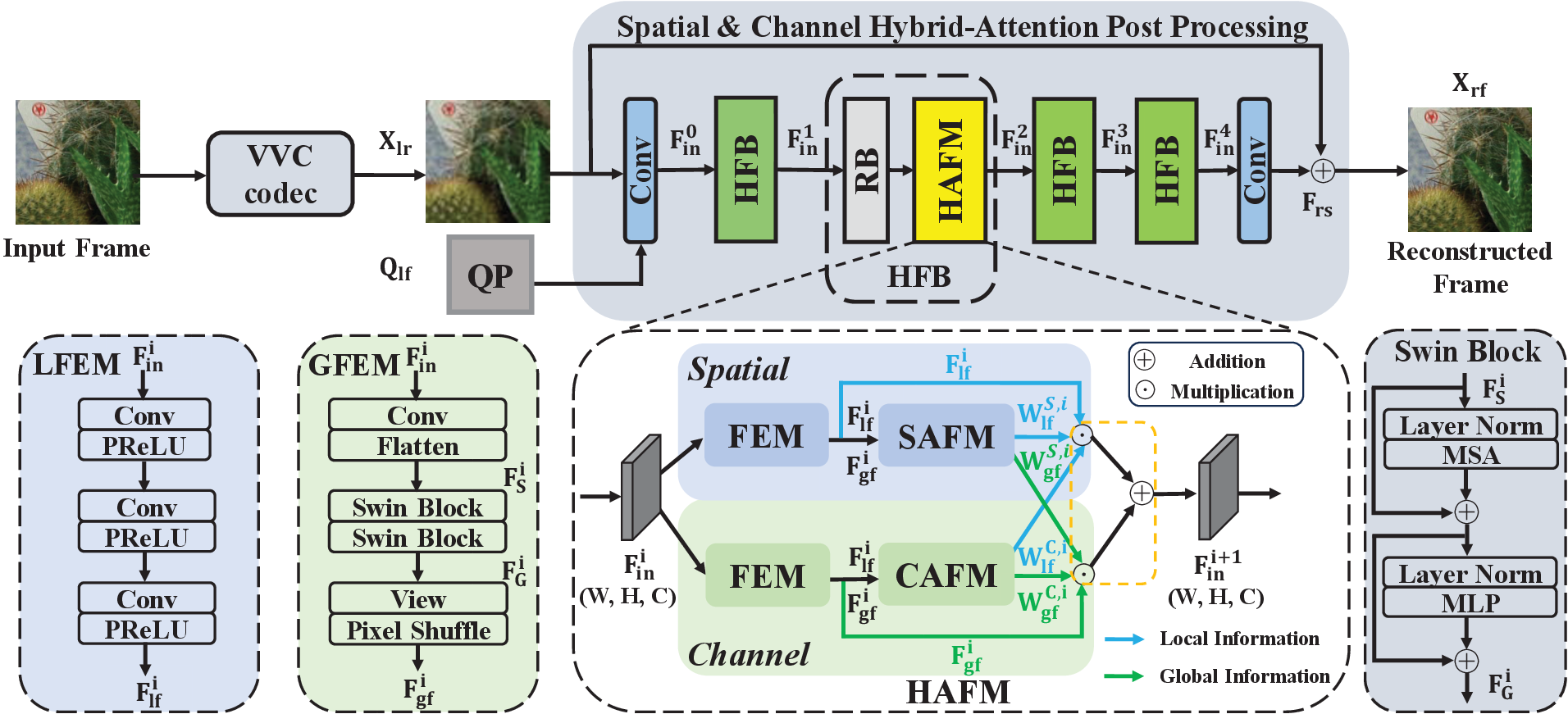}
    \caption{The architecture of the proposed SC-HVPPNet for video post-processing.}
    \label{fig:overview}
\end{figure*}
\textbf{3) CNN and Transformer fusion}-based methods ~\cite{Peng2021ConformerLF, Liu2023LightweightII} are devised to take advantage of both CNN and Transformer, thereby enhancing the representation capability of local features and global representations.
Nonetheless, the integration approaches of current fusion-based methods are relatively monotonous. 
Thus, the essential issue is to explore the interaction between CNN and Transformer.

To address these issues, in this paper, we explore the interaction between CNN and Transformer in the task of video post-processing, and propose a novel Spatial and Channel Hybrid-Attention Video Post-Processing Network (SC-HVPPNet), which can cooperatively exploit the image priors in both spatial and channel domains. 
In the spatial domain, a novel spatial attention fusion module is designed, in which two attention weights are generated to fuse the local and global representations collaboratively. In the channel domain, a novel channel attention fusion module is developed, which can blend the deep representations at the channel dimension dynamically.
Extensive experiments show that SC-HVPPNet notably boosts video restoration quality in the VTM-11.0-NNVC RA configuration.

Compared with existing VPP methods, the contributions of this paper can be summarised below:

\textbf{1)} We propose a novel Spatial and Channel Hybrid-Attention Video Post-Processing Network (SC-HVPPNet), which can cooperatively exploit the image priors in both spatial and channel domains.

\textbf{2)} In the spatial domain, a Spatial Attention Fusion Module (SAFM) is presented to collaboratively fuse the local and global representations by the respective attention weights.

\textbf{3)} In the channel domain, a Channel Attention Fusion Module (CAFM) is developed to dynamically blend the deep representations at the channel dimension.

\textbf{4)} Extensive experiments show that SC-HVPPNet notably boosts video restoration quality, with an average reduction in bitrate of 5.29\%, 12.42\%, and 13.09\% for Y, U, and V components in the VTM-11.0-NNVC RA configuration.

\section{THE PROPOSED METHOD}

\subsection{Overview of SC-HVPPNet}

As illustrated in Fig. \ref{fig:overview}, given the lossy video frame $\mathbf{X}_{lr} \in \mathbb{R}^{W \times H \times C_{in}}$ (\textit{W}, \textit{H} and \textit{$C_{in}$} are
the frame height, width and input channel number), and the corresponding QP value $\mathbf{Q}_{lf}$,
a $3 \times 3$ convolutional layer is employed to fuse them initially.
Meanwhile, the channel dimension is increased and the shallow feature $\mathbf{F}_{in}^{0} \in \mathbb{R}^{W \times H \times C}$ ($C$ is the feature channel number) of the frame is extracted.
Then, several Hybrid Fusion Blocks (HFB) perform deep feature extraction on $\mathbf{F}_{in}^{0}$.
Concretely, the HFB initially preprocesses the shallow feature $\mathbf{F}_{in}^{i}$ using several Residual Blocks (RB), and subsequently, for feature restoration, it employs a Hybrid-Attention Fusion Module (HAFM) that considers feature interaction from both spatial and channel domains.
To recover the residual map $\mathbf{F}_{in}^{4}$ output by these HFBs, a $3 \times 3$ convolutional layer is applied.
Ultimately, the restored residual $\mathbf{F}_{rs}$ is superimposed with the original lossy video frame $\mathbf{X}_{lr}$ through skip connections to obtain the final reconstructed video frame $\mathbf{X}_{rf}$.

\subsection{The Network Architecture}
\label{}
To clearly describe, we first introduce the individual feature interactions within the spatial and channel domains, followed by the cooperation between spatial and channel features. Finally, we detail the specifics of the network architecture.

\begin{figure}
  \centering
    \includegraphics[width=1.0\linewidth]{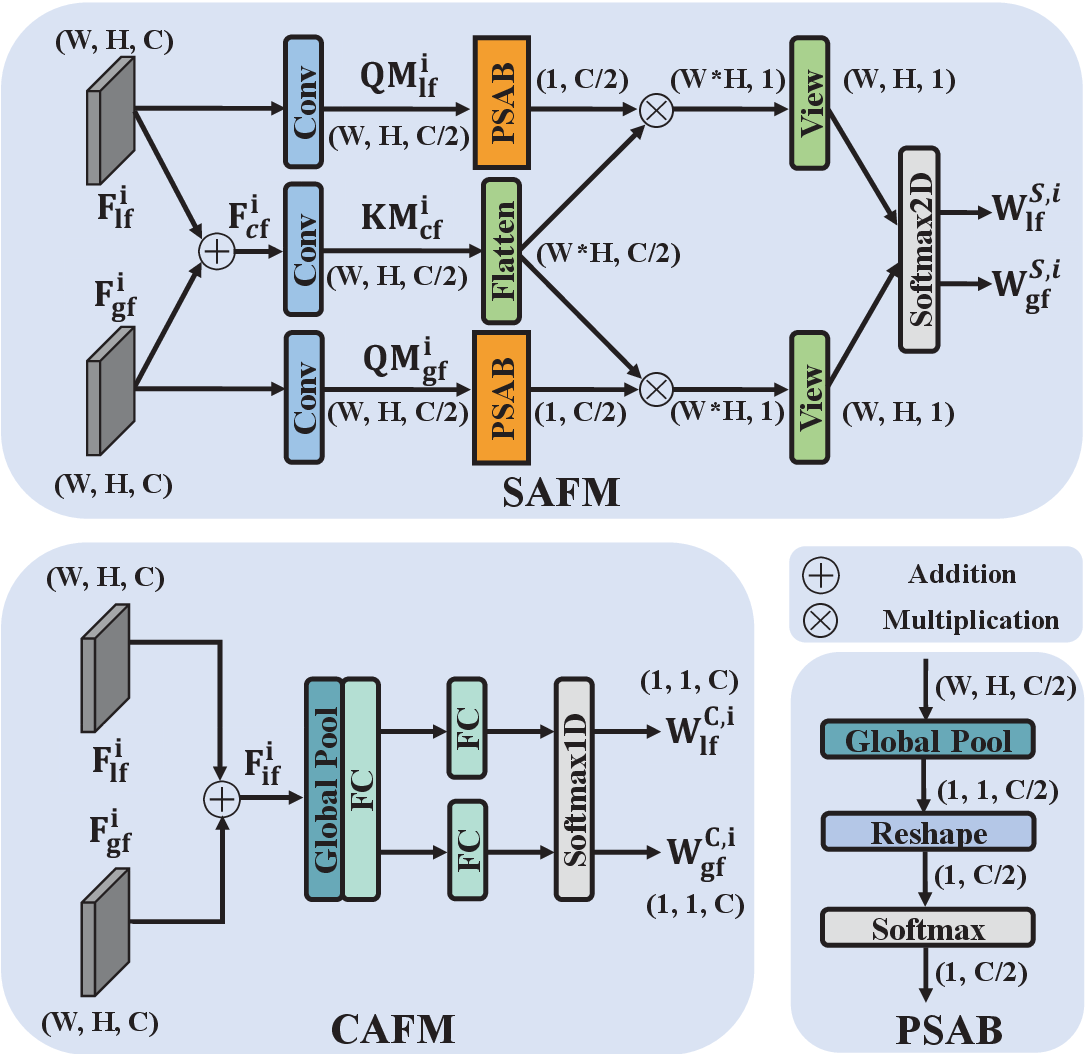}
    \caption{The SAFM and CAFM of SC-HVPPNet.}
    \label{fig:CAFM+SAFM}
\end{figure}

\textbf{Feature interaction in spatial domain:} With this regard, two modules are comprised: the Feature Extraction Module (FEM) and the Spatial Attention Fusion Module (SAFM).
The FEM is responsible for initially processing $\mathbf{F}_{in}^{i}$, which is extracted as $\mathbf{F}_{lf}^{i}$ for local representation and $\mathbf{F}_{gf}^{i}$ for global representation. The specific structure of FEM will be discussed in the latter of this section.
As for the SAFM, the polarized attention and data enhancement concepts of ~\cite{Liu2021PolarizedST} are employed to explore the weight of $\mathbf{F}_{lf}^{i}$ and $\mathbf{F}_{gf}^{i}$ in the spatial domain (represented as $\mathbf{W}_{lf}^{S,i}$ and $\mathbf{W}_{gf}^{S,i}$).
As shown in Fig. \ref{fig:CAFM+SAFM}, SAFM is a three-branch fusion network based on the non-local self-similar structure.
Two convolutional layers are utilized to achieve the query feature maps $\mathbf{QM}_{lf}^{i}, \mathbf{QM}_{gf}^{i} \in \mathbb{R}^{W \times H \times \frac{C}{2}}$ on $\mathbf{F}_{lf}^{i}$ and $\mathbf{F}_{gf}^{i}$.
Moreover, to attain the key feature map $\mathbf{KM}_{cf}^{i} \in \mathbb{R}^{W \times H \times \frac{C}{2}}$, $\mathbf{F}_{lf}^{i}$ and $\mathbf{F}_{gf}^{i}$ are aggregated to a fused feature $\mathbf{F}_{cf}^{i}$ in a roughly way primarily.
Then $\mathbf{F}_{cf}^{i}$ is passed through a
 convolutional layer and is flattened to a size of $(W \times H) \times \frac{C}{2}$ for subsequent weight calculations.
After that, two Polarized Self-Attention Blocks (PSAB) are applied to reduce the computational complexity of the spatial similarity matrix, which applies a global pooling to reduce the dimension of $\mathbf{QM}_{lf}^{i}$ and $\mathbf{QM}_{gf}^{i}$ into a size of $1 \times 1 \times \frac{C}{2}$.
Then, we reshape the feature to a size of $1 \times \frac{C}{2}$ and employ a Softmax function to intensify these two query feature maps.
Crucially, we calculate the correlation of $\mathbf{QM}_{lf}^{i}$ and $\mathbf{QM}_{gf}^{i}$ with $\mathbf{KM}_{cf}^{i}$ correspondingly for each, which represents the weight of $\mathbf{F}_{lf}^{i}$ and $\mathbf{F}_{gf}^{i}$ in $\mathbf{F}_{cf}^{i}$.
Finally, we apply the View and Softmax2D functions to normalize the obtained two weights, yielding the final weights $\mathbf{W}_{lf}^{S,i}$ and $\mathbf{W}_{gf}^{S,i}$.
Consequently, the process of SAFM can be expressed in \eqref{eq}:
\begin{equation}
\begin{split}
\mathbf{W}_{lf}^{S,i} = SM2D(\sigma_{1}\left [ PSAB(\mathbf{QM}_{lf}^{i}) \times \sigma_{2}(\mathbf{KM}_{cf}^{i})\right ]\\
\mathbf{W}_{gf}^{S,i} = SM2D(\sigma_{1}\left [ PSAB(\mathbf{QM}_{gf}^{i}) \times \sigma_{2}(\mathbf{KM}_{cf}^{i})\right ]\\
\end{split}
\label{eq}
\end{equation}
where $SM2D$ is the Softmax2D function, $\sigma_{1}$ is the View function, $\sigma_{2}$ is the Flatten function and ”$\times$” is the matrix dot-product operation. 

\textbf{Feature interaction in channel domain:}
Similar to the spatial domain, the feature interaction in the channel domain also contains the same FEM and the corresponding Channel Attention Fusion Module (CAFM).
The CAFM is applied to excavate the weight of $\mathbf{F}_{lf}^{i}$ and $\mathbf{F}_{gf}^{i}$ in the channel domain (represented as $\mathbf{W}_{lf}^{C,i}$ and $\mathbf{W}_{gf}^{C,i}$).
As shown in Fig. \ref{fig:CAFM+SAFM}, a initially fusion feature $\mathbf{F}_{if}^{i}$ is aggregated by $\mathbf{F}_{lf}^{i}$ and $\mathbf{F}_{gf}^{i}$ firstly.
To shift information from the recovery domain to the weight domain, $\mathbf{F}_{if}^{i}$ is passed to a global average pooling followed by a fully connected layer.
Finally, after passing through a fully connected layer and a softmax function by each, $\mathbf{W}_{lf}^{C,i}$ and $\mathbf{W}_{gf}^{C,i}$ are computed.
Therefore, the mechanism of CAFM is formulated in \eqref{equ:WCF}:
\begin{equation}
\mathbf{W}_{lf}^{C,i},\mathbf{W}_{gf}^{C,i} = SM(FC(GP(\mathbf{F}_{lf}^{i} + \mathbf{F}_{gf}^{i})))
\label{equ:WCF}
\end{equation}
where $SM$ represents the Softmax1D function, $FC$ represents the fully connected layer and $GP$ performs the GlobalPool operation.

\textbf{Feature interaction of spatial and channel domains:}
As enclosed by the orange dashed line in Fig. \ref{fig:overview}, $\mathbf{F}_{lf}^{i}$ and $\mathbf{F}_{gf}^{i}$ are sequentially weighted through channel fusion weights and spatial fusion weights, ultimately leading to the feature interaction in both spatial and channel domains.
Notably, the blue and green lines represent the local information flow and global information flow, individually.
Hence, the deep fusion feature $\mathbf{F}_{in}^{i+1}$ can be formulated in \eqref{eq2}: 
\small
\begin{equation}
\begin{aligned}
\mathbf{F}_{in}^{i+1} &= \mathbf{W}_{lf}^{C,i} \odot (\mathbf{W}_{lf}^{S,i} \odot \mathbf{F}_{lf}^{i})
             + \mathbf{W}_{gf}^{C,i} \odot (\mathbf{W}_{gf}^{S,i} \odot \mathbf{F}_{gf}^{i}) \\
             &= (\mathbf{W}_{lf}^{C,i} \odot \mathbf{W}_{lf}^{S,i}) \odot \mathbf{F}_{lf}^{i}
             + (\mathbf{W}_{gf}^{C,i} \odot \mathbf{W}_{gf}^{S,i}) \odot \mathbf{F}_{gf}^{i} \\
             &= \mathbf{W}_{lf}^{CS,i} \odot \mathbf{F}_{lf}^{i} + \mathbf{W}_{gf}^{CS,i} \odot \mathbf{F}_{gf}^{i} \\    
\end{aligned}
\label{eq2}
\end{equation}
\normalsize
where $\mathbf{W}_{lf}^{CS,i}, \mathbf{W}_{gf}^{CS,i} \in \mathbb{R}^{W \times H \times C}$ are the deep fusion weights which take both space and channel domains into account and $\odot$ is the multiplication operation.

\textbf{Other network details:}
The arrangement of RB in HFB involves a skip connection consisting of two $3 \times 3$ convolutional layers with 1 PReLU.
Moreover, the FEM in both spatial and channel domains is composed of a Local Feature Extraction Module (LFEM) based on CNN and a Global Feature Extraction Module (GFEM) based on Swin-Transformer, respectively.
As detailed in Fig. \ref{fig:overview}, LFEM applies several cascaded $3 \times 3$ convolutional layers with PReLU.
To bridge the gap between the 3D features of CNN and the 2D format of Transformer, GFEM uses a convolutional layer with a $4 \times 4$ kernel to downsample the $\mathbf{F}_{in}^{i}$ into the size of $\frac{WH}{4^2} \times {4^2} \times C$. 
And the split feature is flattened to $\mathbf{F}_{S}^{i} \in \mathbb{R}^{(\frac{W}{4} \times \frac{H}{4}) \times {4^2} \times C}$ for subsequent Swin Blocks.
Specifically, the Swin Blocks use MSA and MLP for calculating the global similarity and the nonlinear transformation.
To align the shape of $\mathbf{F}_{G}^{i}$ with that of $\mathbf{F}_{in}^{i}$, 
we reshape the 2D feature $\mathbf{F}_{G}^{i}$ to a 3D feature ($\frac{W}{4} \times \frac{H}{4} \times {4^2} \times C$ ) by the View function.
A PixelShuffle layer is utilized to upsample the feature to a size of $W \times H \times C$ afterward.
Finally, $\mathbf{F}_{gf}^{i}$ is obtained.

\begin{table*}[!t]
  \small
  \centering
  \caption{Performance comparison of SC-HVPPNet compared with VTM-11.0-NNVC under RA configuration (GPU).}
  \setlength{\tabcolsep}{7.0mm}{
  \begin{tabular}{ccccccc}
    \toprule
     & Y-PSNR & U-PSNR & V-PSNR & Y-MSIM & U-MSIM & V-MSIM\\
    \midrule
    Class A1 & -5.10\% & -8.66\% & -9.46\% & -5.94\% & -12.83\% & -13.15\%\\
    Class A2 & -6.12\% & -13.89\% & -8.66\% & -5.72\% & -14.16\% & -6.91\% \\
    Class B & -3.00\% & -11.85\% & -12.16\% & -4.64\% & -15.77\% & -15.31\% \\
    Class C & -5.49\% & -16.90\% & -17.57\% & -4.93\% & -17.71\% & -16.74\% \\
    Class D & -8.66\% & -18.74\% & -21.61\% & -4.88\% & -17.83\% & -17.82\% \\
    Overall & -5.54\% & -14.18\% & -14.31\% & -5.13\% & -15.89\% & -14.47\% \\
    \bottomrule
  \end{tabular}}
  \label{table:compression}
\end{table*}
\begin{table*}[!t]
  \small
  \centering
  \caption{The BD-rate results of PSNR for SC-HVPPNet over VTM-11.0-NNVC under RA configuration at different QPs.}
  \setlength{\tabcolsep}{5.2mm}{
  \begin{tabular}{|c|c|c|c|c|c|c|c|}
    \hline
    \multirow{2}{*}{Class} & \multirow{2}{*}{Sequence} & \multicolumn{3}{c|}{Low QP(22-37)} & \multicolumn{3}{c|}{High QP(27-42)}\\
    \cline{3-8}
    &  &  Y & U & V & Y & U & V\\
    \hline
    \multirow{3}{*}{A1} & Tango2 & -5.83\% & -23.45\% & -16.19\% & -6.31\% & -24.55\% & -16.53\%\\
   \multirow{3}{*}{} & FoodMarket4 & -4.68\% & -5.51\% & -5.17\% & -5.36\% & -6.87\% & -6.72\%\\
    \multirow{3}{*}{} & Campfire & -3.50\% & 6.28\% & -4.86\% & -4.82\% & 0.62\% & -7.61\%\\
    \hline
    \multirow{3}{*}{A2} & CatRobot & -7.12\% & -19.21\% & -13.93\% & -7.37\% & -21.58\% & -15.29\%\\
    \multirow{3}{*}{} & DaylightRoad2 & -9.36\% & -18.12\% & -9.85\% & -8.46\% & -21.31\% & -10.23\%\\
    \multirow{3}{*}{} & ParkRunning3 & -2.14\% & 0.36\% & 0.46\% & -2.51\% & -2.71\% & -2.97\%\\
    \hline
    \multirow{5}{*}{B} & MarketPlace & -4.25\% & -19.80\% & -16.81\% & -4.50\% & -23.14\% & -19.95\%\\
    \multirow{5}{*}{} & RitualDance & -5.60\% & -12.32\% & -17.50\% & -5.49\% & -14.46\% & -20.09\%\\
    \multirow{5}{*}{} & Cactus & -1.81\% & 5.70\% & 1.90\% & -3.71\% & -3.78\% & -4.55\%\\
    \multirow{5}{*}{} & BasketballDrive & --4.68\% & -11.66\% & -13.23\% & -5.17\% & -13.56\% & -15.02\%\\
    \multirow{5}{*}{} & BQTerrace & 5.23\% & -9.75\% & -5.47\% & -0.73\% & -12.68\% & -11.02\%\\
    \hline
    \multirow{4}{*}{C} & BasketballDrill & -5.33\% & -15.65\% & -14.58\% & -5.41\% & -20.41\% & -18.15\%\\
    \multirow{4}{*}{} & BQMall & -5.93\% & -16.98\% & -19.81\% & -6.56\% & -20.93\% & -23.52\%\\
    \multirow{4}{*}{} & PartyScene & -7.57\% & -11.56\% & -11.93\% & -7.71\% & -16.19\% & -15.59\%\\
    \multirow{4}{*}{} & RaceHorses & -2.52\% & -14.99\% & -17.96\% & -3.88\% & -20.36\% & -21.83\%\\
    \hline
    \multirow{4}{*}{D} & BasketballPass & -8.20\% & -22.08\% & -23.58\% & -8.47\% & -25.90\% & -24.94\%\\
    \multirow{4}{*}{} & BQSquare & -14.05\% & -10.95\% & -22.96\% & -13.88\% & -12.35\% & -25.29\%\\
    \multirow{4}{*}{} & BlowingBubbles & -6.54\% & -13.82\% & -12.98\% & -6.89\% & -17.12\% & -15.55\%\\
    \multirow{4}{*}{} & RaceHorses & -6.57\% & -22.50\% & -24.23\% & -6.46\% & -27.21\% & -26.98\%\\
    \hline
    \multicolumn{2}{|c|}{\textbf{Average}} & \textbf{-5.29\%} & \textbf{-12.42\%} & \textbf{-13.09\%} & \textbf{-5.98\%} & \textbf{-16.03\%} & \textbf{-15.89\%}\\
    \hline
  \end{tabular}}
  \label{table:2}
\end{table*}

\subsection{Loss Function}
We employ the CharbonnierLoss ~\cite{413553} as the loss function, given the initial lossless frame $\mathbf{X}$ and the final recovered frame $\mathbf{\hat{X}}$, the loss is defined in \eqref{eq3}:
\begin{equation}
\mathcal{L}(\mathbf{X}, \mathbf{\hat{X}}) = \frac{1}{N} \sum_{i=1}^{N} \sqrt{(\mathbf{X} - \mathbf{\hat{X}})^2 + \epsilon^2} 
\label{eq3}
\end{equation}
where $N$ is the number of samples used in each training iteration and $\epsilon$ is a small positive number.
Considering the human eye is less sensitive to color details and more focused on brightness in images, the components of YUV are weighted separately, and the final loss function is in \eqref{eq4}:
\begin{equation}
\mathcal{L} = 10 * \mathcal{L}(\mathbf{X}_Y, \mathbf{\hat{X}}_Y) + \mathcal{L}(\mathbf{X}_U, \mathbf{\hat{X}}_U) + \mathcal{L}(\mathbf{X}_V, \mathbf{\hat{X}}_V) 
\label{eq4}
\end{equation}
with the specific weighting ratio being Y: U: V = 10: 1: 1.

\section{EXPERIMENTS AND RESULTS}
\subsection{Implementation and Training Details}
In this paper, we select 191 video sequences with YCbCr 4:2:0 format from each resolution of BVI-DVC ~\cite{Ma2022BVIDVCAT} for training according to common test conditions (CTC).
All sequences are compressed encoded by VTM-11.0-NNVC using the JVET-CTC ~\cite{unknown} under RA configuration with five QP values: 22, 27, 32, 37 and 42. 
\begin{figure}
  \centering
    \includegraphics[width=1.0\linewidth]{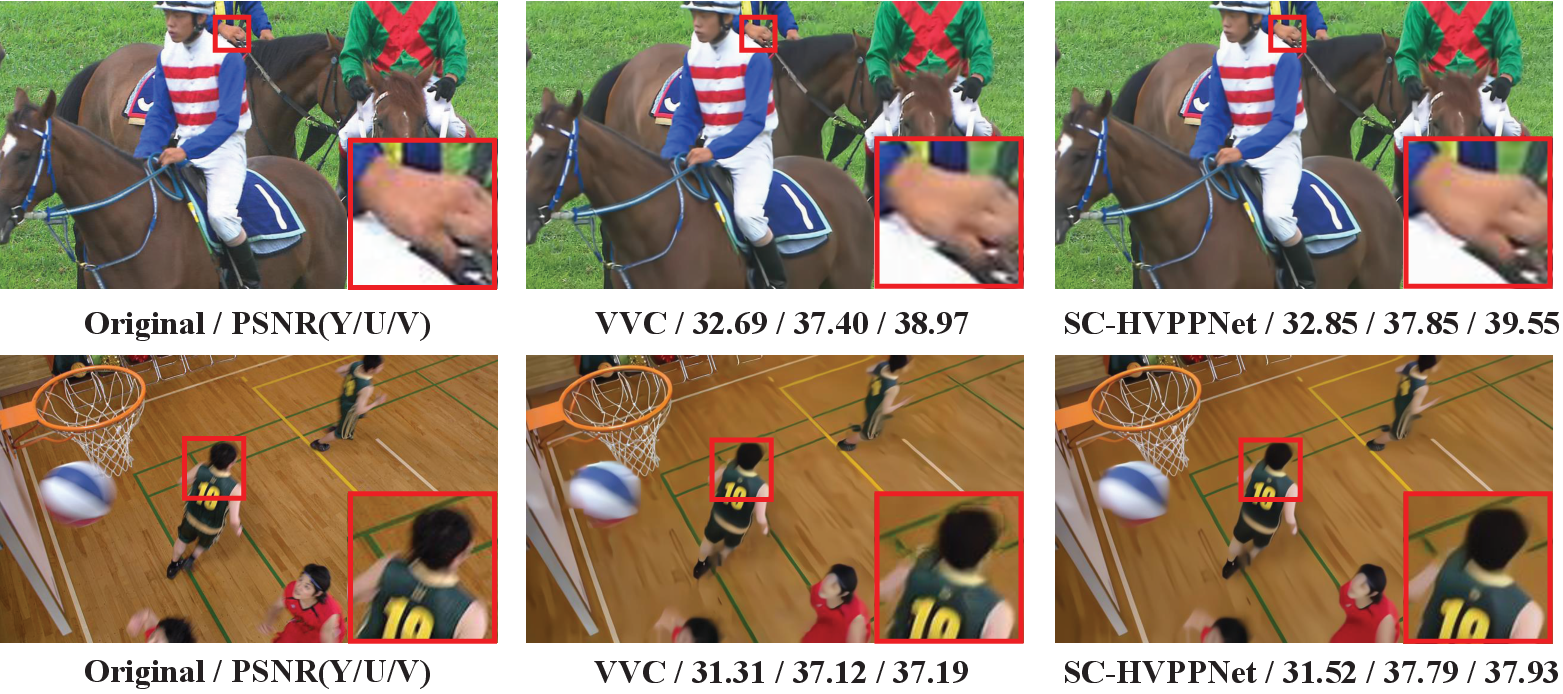}
    \caption{Visual comparison on the 4-th frame of \textit{RaceHorses} and \textit{BasketballDrill} when QP=32 and QP=42.}
    \label{fig:visual}
\end{figure}
After that, the output code stream is decoded to obtain the compressed video sequence.
Our SC-HVPPNet is trained on an NVIDIA GeForce RTX 3060 GPU by PyTorch.
We train our network for 31 epochs and $1 \times 10^6$ iterations total using Adam with $\beta_{1} = 0.9$, $\beta_{2} = 0.999$.
The learning rate is initialized as $2 \times 10^{-4}$, which is reduced by half after every $10^5$ iteration. 
We set the CNN layer number of LFEM as 3.
\begin{table*}[!t]
  \small
  \centering
  \caption{The BD-rate results of different VPP models in the low QP range (22-37) for Y-PSNR under RA configuration. The parentheses indicate the version of the VVC codec framework corresponding to the model used during the experiments.}
  \setlength{\tabcolsep}{4.7mm}{
  \begin{tabular}{ccccccc}
    \toprule
    Method & A1 & A2 & B & C & D & Overall\\
    \midrule
    VTM-PP (VTM-4.0) ~\cite{9102912} & -2.41\% & -4.22\% & -2.57\% & -3.89\% & -5.80\% & -3.76\% \\       
    JVET-O0079 (VTM-5.0) ~\cite{8852743} & -0.87\% & -1.68\% & -1.47\% & -3.34\% & -4.97\% & -2.47\% \\
    JVET-T0079 (VTM-10.0) ~\cite{9856937} & -2.86\% & -2.98\% & -2.92\% & -2.96\% & -3.48\% & -3.04\% \\
    WCDANN (VTM-11.0-NNVC) ~\cite{10201857} & -2.23\% & -2.70\% & -2.73\% & -3.43\% & -4.76\% & -2.81\% \\ 
    JVET-Z0144 (VTM-11.0-NNVC) ~\cite{JVET04} & -2.14\% & -2.56\% & -2.54\% & -3.33\% & -4.54\% & -3.07\% \\
    JVET-Z0082 (VTM-11.0-NNVC) ~\cite{JVET03} & \textbf{-6.88\%} & -4.52\% & \textbf{-5.56\%} & -3.94\% & -4.96\% & -5.14\% \\ 
    \textbf{SC-HVPPNet (VTM-11.0-NNVC) (ours)} & -4.67\% & \textbf{-6.21\%} & -2.22\% & \textbf{-5.34\%} & \textbf{-8.84\%} & \textbf{-5.29\%} \\
    \bottomrule
  \end{tabular}}
  \label{table:other}
\end{table*}
\begin{table*}[htbp]
  \small
  \centering
  \caption{Ablation study of SAFM and CAFM in terms of BD-Rate on Y-PSNR in the low QP range over VTM-11.0-NNVC.}
  \setlength{\tabcolsep}{7.4mm}{
  \begin{tabular}{ccccccc}
    \toprule
    Method & A1 & A2 & B & C & D & Overall\\
    \midrule
    S-HVPPNet & -2.17\% & -2.96\% & -1.63\% & -3.44\% & -6.61\% & -3.35\% \\
    C-HVPPNet & -0.63\% & -1.51\% & -0.03\% & -1.89\% & -4.17\% & -1.62\% \\   
    SC-HVPPNet & \textbf{-4.67\%} & \textbf{-6.21\%} & \textbf{-2.22\%} & \textbf{-5.34\%} & \textbf{-8.84\%} & \textbf{-5.29\%} \\
    \bottomrule
  \end{tabular}}
  \label{table:ablation}
\end{table*}
\begin{table}[htbp]
  \small
  \centering
  \caption{Ablation study of fusion manners.}
  \setlength{\tabcolsep}{3.4mm}{
  \begin{tabular}{cccc}
    \toprule
    Method & SQ-HVPPNet & P-HVPPNet & SC-HVPPNet\\
    \midrule
    Overall & -5.02\% & -4.87\% & \textbf{-5.29\%} \\
    \bottomrule
  \end{tabular}
  }
  \label{table:manner}
\end{table}
Note that only one network model is trained for all QP modes.
During training, the video frames are randomly segmented into $256 \times 256$ image blocks, and converted to YCbCr 4:4:4 format.
For the testing session, we deploy the standard test sequences from JVET-CTC.

\subsection{Comparison with Other VPP Methods}\label{AA}
According to JVET-CTC, Table. \ref{table:compression} illustrates the overall bitrate savings of the proposed SC-HVPPNet according to PSNR and MS-SSIM (MSIM) compared to VTM-11.0-NNVC under five QP values (22, 27, 32, 37, 42).
It can be observed that SC-HVPPNet can greatly save the required bitrate under the same quality effect, with average BD-rates of YUV components -5.54\%, -14.18\%, and -14.31\% on PSNR, respectively.
And the results on the chroma component are much higher than that on the luminance component.
As shown in Table. \ref{table:2}, the coding gains in the higher QP ranges consistently outperform those in the lower QP ranges, especially in the chroma components, where the gains in higher QPs are 3.61\% and 2.80\% greater than those in the low QP range.
As depicted in Fig. \ref{fig:visual}, the ringing effect of the hand fringe areas in \textit{RaceHorses} and the blocking artifact and uneven edge pixels around the head in \textit{BasketballDrill} are improved.
Moreover, we compare SC-HVPPNet with 6 state-of-the-art VPP methods developed for the VVC RA configuration. 
These include VTM-PP ~\cite{9102912}, JVET-O0079 ~\cite{8852743}, JVET-T0079 ~\cite{9856937}, WCDANN ~\cite{10201857}, JVET-Z0144 ~\cite{JVET04} and JVET-Z0082 ~\cite{JVET03}.
As shown in Table. \ref{table:other}, the proposed SC-HVPPNet offers the best performance when compared to the other 6 methods, especially in lower-resolution videos, such as Class C and Class D.
\begin{figure}[htb]
\begin{minipage}[b]{1.0\linewidth}
	\centerline{\includegraphics[width=3.2in]{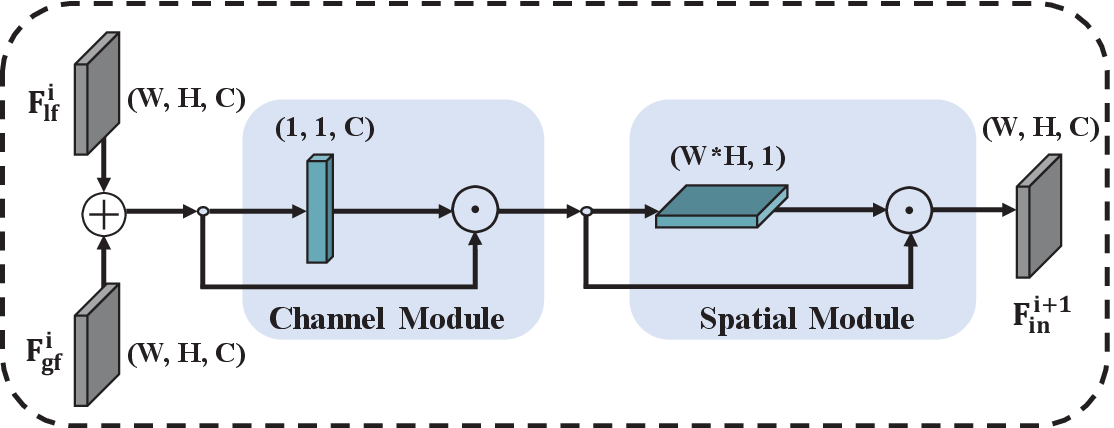}}
        \centerline{(a)}
        \label{fig:sequential}
\end{minipage}
\begin{minipage}{0.49\linewidth}
	\centerline{\includegraphics[width=1.5in]{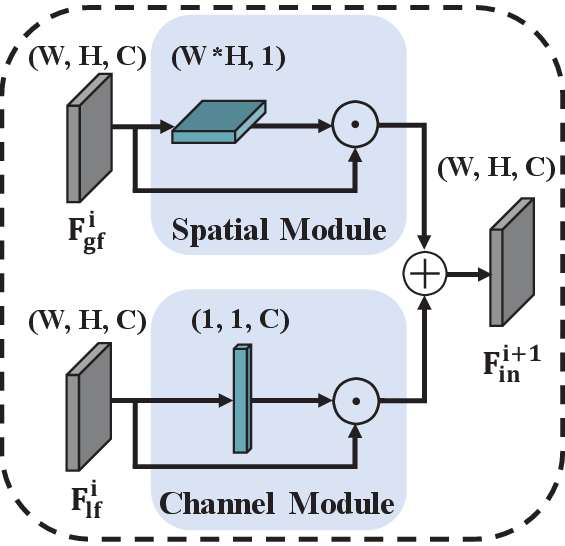}}   % 换行
        \centerline{(b)}
        \label{fig:parallel}
\end{minipage} % 中间不空行代表不换行
\begin{minipage}{0.49\linewidth}
	\centerline{\includegraphics[width=1.5in]{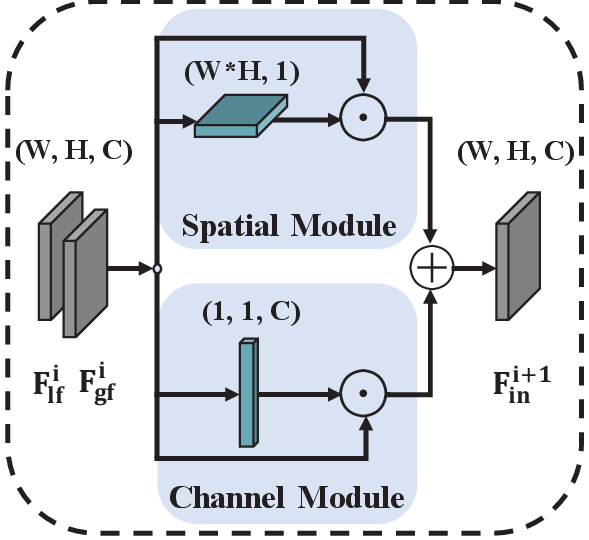}}
        \centerline{(c)}
        \label{fig:c}
\end{minipage} % 中间不空行代表不换行
\caption{The architectures of different fusion manners. a) sequential, b) parallel, c) hybrid interaction. $\odot$ is the multiplication operator and $\oplus$ is the element-wise addition.}
\label{fig:submodules}
\end{figure}
Furthermore, as SC-HVPPNet trains a single model for all sequences and codec scenarios, it eliminates the requirement for model selection or extra coding parameters.

\subsection{Ablation Study}
To analyze the impact of interaction manners between spatial and channel features, we explore three fusion manners for the spatial and channel modules of SC-HVPPNet: 
\textbf{a)} fusion in sequential, \textbf{b)} fusion in parallel, and \textbf{c)} fusion in hybrid interaction, as shown in Fig. \ref{fig:submodules}.
The spatial and channel modules correspond to SAFM and CAFM in Fig. \ref{fig:overview}, respectively.
In the sequential approach, fused features of $\mathbf{F}_{lf}^{i}$ and $\mathbf{F}_{gf}^{i}$ pass through channel and spatial modules in order, while spatial and channel features calculate their respective attentions in a parallel way, which are then fused.
We argue that spatial and channel attention are crucial in the restoration process for local and global features.
Hence, we fuse $\mathbf{F}_{lf}^{i}$ and $\mathbf{F}_{gf}^{i}$ in a more fine-grained hybrid interaction manner in the spatial and channel domains, which is deployed in our proposed SC-HVPPNet.
Moreover, we integrate the first two fusion methods into SC-HVPPNet, named SQ-HVPPNet and P-HVPPNet.
The overall BD-rate results for PSNR on Y components in the low QP range are shown in Table.\ref{table:manner}. It can be observed that SC-HVPPNet offers the best performance compared to SQ-HVPPNet and P-HVPPNet.
In addition, we retrain our SC-HVPPNet by using pure spatial (SAFM) and channel (CAFM) attention fusions, respectively (regarded as S-HVPPNet and C-HVPPNet). 
As shown in Table.\ref{table:ablation}, SC-HVPPNet increases the gain with an overall BD-rate of 1.93\% and 3.66\% compared with S-HVPPNet and C-HVPPNet.
We conclude that SC-HVPPNet performs attractively by combining spatial and channel attention mechanisms.

\section{CONCLUSION}
In this paper, we explore the interaction between CNN and Transformer in the task of video post-processing, and propose a novel Spatial and Channel Hybrid-Attention Video Post-Processing Network (SC-HVPPNet), which can cooperatively exploit the image priors in both spatial and channel domains.
The designed hybrid-attention structures enhance both local and global modeling capabilities in an efficient feature interaction way.
Extensive experiments show that our SC-HVPPNet notably boosts video restoration quality in the VTM-11.0-NNVC RA configuration.

\section*{Acknowledgment}
This work was supported in part by the Jiangsu Provincial Science and Technology Plan (BE2021086), the National Natural Science Foundation of China (NSFC) under grants 62302128 and 62076080, and the Postdoctoral Science Foundation of Heilongjiang Province of China (LBH-Z22175).
\bibliographystyle{plain} 
\bibliography{refs}

\end{document}